\documentclass[preprint,10pt]{elsarticle}

\usepackage{textcomp,booktabs}
\usepackage[usenames,dvipsnames]{color}
\usepackage{colortbl}
\definecolor{mygray}{gray}{.9}
\definecolor{mypink}{rgb}{.99,.91,.95}
\definecolor{mycyan}{cmyk}{.3,0,0,0}
\usepackage{amssymb}
\usepackage{graphicx}
\usepackage{booktabs}
\usepackage{tabularx}
\usepackage{array}
\usepackage{lscape}
\usepackage{longtable}
\usepackage{multirow}
\usepackage{amsmath}
\usepackage{colortbl}
\usepackage{epsfig}
\usepackage{epsf}
\usepackage{subfigure}
\usepackage{rotating}
\usepackage{pdflscape}
\usepackage{lscape}
\usepackage{indentfirst}
\usepackage{wrapfig}
\usepackage{CJK}
\usepackage[amsmath,thmmarks]{ntheorem}
\biboptions{numbers,sort&compress}

\newcommand{\PreserveBackslash}[1]{\let\temp=\\#1\let\\=\temp}
\newcolumntype{C}[1]{>{\PreserveBackslash\centering}p{#1}}
\newcolumntype{R}[1]{>{\PreserveBackslash\raggedleft}p{#1}}
\newcolumntype{L}[1]{>{\PreserveBackslash\raggedright}p{#1}}
\newtheorem{definition}{Definition}[section]

\begin{document}

\begin{frontmatter}

\title{Multi-sensor data fusion based on a generalised belief divergence measure}

\author[address1]{Fuyuan Xiao\corref{label1}}
\ead{xiaofuyaun@swu.edu.cn}
\address[address1]{School of Computer and Information Science, Southwest University, Chongqing, 400715, China}
\cortext[label1]{Corresponding author at: School of Computer and Information Science, Southwest University, No.2 Tiansheng Road, BeiBei District, Chongqing, 400715, China. %E-mail: xiaofuyuan@swu.edu.cn
}

\begin{abstract}
Multi-sensor data fusion technology plays an important role in real applications.
Because of the flexibility and effectiveness in modelling and processing the uncertain information regardless of prior probabilities, Dempster-Shafer evidence theory is widely applied in a variety of fields of information fusion.
However, counter-intuitive results may come out when fusing the highly conflicting evidences.
In order to deal with this problem, a novel method for multi-sensor data fusion based on a new generalised belief divergence measure of evidences is proposed.
Firstly, the reliability weights of evidences are determined by considering the sufficiency and importance of the evidences.
After that, on account of the reliability weights of evidences, a new Generalised Belief Jensen-Shannon divergence (GBJS) is designed to measure the discrepancy and conflict degree among multiple evidences, which can be utilised to measure the support degrees of evidences.
Afterwards, the support degrees of evidences are used to adjust the bodies of the evidences before using the Dempster's combination rule.
Finally, an application in fault diagnosis demonstrates the validity of the proposed method.
%The results show that the proposed method outperforms other related methods where the basic belief assignment (BBA) of the true target is 89.73\%.
\end{abstract}

\begin{keyword}
Sensor data fusion, Dempster--Shafer evidence theory, Evidential conflict, Generalised belief divergence measure, Generalised Jensen--Shannon divergence, Fault diagnosis
\end{keyword}

\end{frontmatter}

\section{Introduction}\label{Introduction}
With the fast development of electronic technologies, a variety of sensors were developed and applied in many engineering fields, like the fault diagnosis~\cite{hang2014fault,jiang2016sensordiagnosis,cheng2016new,geng2017model},
wireless sensor networks~\cite{zhang2014green,zhang2015belief,zhang2016toward},
risk analysis~\cite{vandecasteele2016reasoning},
and so on~\cite{dong2007hidden,yang2013fusion,shen2014sensor}.
Because multi-sensor-based applications can provide more reliable and accurate information than that of a single sensor alone, multi-sensor data fusion technologies have attracted considerable attentions in many fields of practical applications for the past few years~\cite{he2005Survey,he2000Multisensor,he1996Survey}.
However, due to the influence of the environment, such as, bad weather conditions, sensor failures, wireless communication problems, the uncertainty and imprecision are unavoidable in the course of sensor data collection.
How to model and copy with these kinds of uncertain information is still an open question~\cite{Liu2017Ordered}.
To address this question, various theories were presented for multi-sensor data fusion,
consisting of the rough sets theory~\cite{walczak1999rough,shen2000fault},
fuzzy sets theory~\cite{zadeh1965fuzzy,liu2013fuzzy,mardani2015fuzzy,jiang2018IJSS,zhang2017multiple},
evidence theory~\cite{Dempster1967Upper,Shafer1978A,deng2016evidence,jiang2017MPE,zhang2014novelParadigm},
evidential reasoning~\cite{yang2013evidential,yang2014interactive,Fu2012Anevidential,Fu2011Anattribute},
D numbers theory~\cite{Bian2018Failure,xiao2016intelligent},
Z numbers~\cite{zadeh2011note,Kang2017Stable},
and so on~\cite{ma2012qualitative,XU2017AMM,Fu2016Distributed}.

Dempster--Shafer (D--S) evidence theory, which was firstly proposed by Dempster~\cite{Dempster1967Upper} and was extended by Shafer~\cite{Shafer1978A} is an useful uncertainty reasoning approach.
D--S evidence theory is flexible and effective in modelling both of the uncertainty and imprecision without prior information, so that it has been applied broadly in all kinds of fields, such as fault diagnosis~\cite{Jiang2017FMEA,fan2006fault,yuan2016modeling,xiao2017faultdiagnosis},
decision making~\cite{jiang2017Intuitionistic,Xiao2017AnImproved,Fu2017Determining,Wang2017IJCCC},
pattern recognition~\cite{denoeux1995k,ma2016evidential,liu2016adaptive},
risk analysis~\cite{dutta2015uncertainty},
supplier selection~\cite{liuDEMATEL2017},
human reliability analysis~\cite{zhengxianglin2017},
and so on~\cite{jiang2018Improved}.
Whereas, when fusing the highly conflicting evidences, the system based on the classical D--S evidence theory may result in the counter-intuitive results~\cite{zadeh1986simple}.
To resolve this issue, many approaches have been presented, which are mainly divided into two categories~\cite{lefevre2002belief,XDWJIJIS21929,han2011weighted}.
One is to revise the Dempster's combination rule, including Smets's unnormalized combination rule~\cite{smets1990combination}, Yager's combination rule~\cite{yager1987dempster}, Dubois and Prade's disjunctive combination rule~\cite{dubois1988representation}, etc.
The other is to pretreat the bodies of evidences, including Murphy's simple average approach~\cite{murphy2000combining}, Deng et al.'s weighted average method~\cite{yong2004combining}, Zhang et al.'s cosine theorem-based method~\cite{zhang2014novel}, Yuan et al.'s entropy-based method~\cite{yuan2016conflict}, etc.
Nevertheless, when revising the Dempster's combination rule, some good properties, like the commutativity and associativity are often destructed.
In addition, if the counter-intuitive results are resulted from the failure of the sensor, the revision of the Dempster's combination rule does not work.
As a result, many research efforts have tended to deal with the bodies of evidences in advance to solve the problem of combining highly conflicting evidences.
In this paper, we also focus on pre-treating the bodies of evidences.
By studying and analysing the existing methods, we found that there still exists some room for the improvement to achieve more precise fusion results.

In this paper, therefore, a novel generalised belief divergence method is first presented to measure the conflict and discrepancy degree among multi-evidences.
Based on that, a new multi-sensor data fusion approach is proposed, which mainly consists of the following three parts.
Firstly, the reliability weights of evidences are determined by considering the sufficiency and importance of the evidences.
Next, on account of the reliability weights of evidences, the presented Generalised Belief Jensen-Shannon divergence is utilised to measure the support degrees of evidences.
Lastly, the support degrees of the evidences are used to adjust the bodies of evidences before using the Dempster's combination rule.
As mentioned above, it can be easy see that the proposed method is concise.
Thanks to taking into account the global conflict and discrepancy degree among multi-evidences, the proposed method is more effective and superior than the existing approaches, which is validated by a numerical example and an application in a motor rotor fault diagnosis.

The rest of this paper is organized as follows.
Section~\ref{Preliminaries} briefly introduces the preliminaries of this paper.
A new Belief Jensen-Shannon divergence is proposed for measuring the distance between the bodies of the evidences in Section~\ref{BJSdivergence}.
A novel multi-sensor data fusion method which is based on the belief divergence measure of evidences and the belief entropy is proposed in Section~\ref{Proposed method}.
%Section~\ref{Experiments} illustrates a numerical example to show the effectiveness of the proposed method.
In Section~\ref{Application}, the proposed method is applied to an application in fault diagnosis.
Finally, Section~\ref{Conclusion} gives a conclusion.

\section{Preliminaries}\label{Preliminaries}

\subsection{Dempster--Shafer evidence theory}
Dempster--Shafer evidence theory~\cite{Dempster1967Upper,Shafer1978A} is applied to deal with uncertain information, belonging to the category of artificial intelligence.
Because of the flexibility and effectiveness in modelling both of the uncertainty and imprecision without prior information, Dempster--Shafer evidence theory requires more weaker conditions than the Bayesian theory of probability.
When the probability is confirmed, Dempster--Shafer evidence theory could convert into Bayesian theory, so it is considered as an extension of the Bayesian theory.
Dempster--Shafer evidence theory has the advantage that it can directly express the ``uncertainty'' by allocating the probability into the subsets of the set which consists multiple objects, rather than to an individual object.
Furthermore, it is capable of combining the bodies of evidences to derive a new evidence.
The basic concepts are introduced as below.

\begin{definition}(Frame of discernment).

Let $U$ be a set of mutually exclusive and collectively exhaustive events, indicted by
\begin{equation}\label{eq_Frameofdiscernment1}
 U = \{E_{1}, E_{2}, \ldots, E_{i}, \ldots, E_{N}\}.
\end{equation}

The set $U$ is called a frame of discernment.
The power set of $U$ is indicated by $2^{U}$, where
\begin{equation}\label{eq_Frameofdiscernment2}
 2^{U} = \{\emptyset, \{E_{1}\}, \ldots, \{E_{N}\}, \{E_{1}, E_{2}\}, \ldots, \{E_{1}, E_{2}, \ldots, E_{i}\}, \ldots, U\},
\end{equation}
and $\emptyset$ is an empty set.
If $A \in 2^{U}$, $A$ is called a proposition.
\end{definition}

\begin{definition}(Mass function).

For a frame of discernment $U$, a mass function is a mapping $m$ from $2^{U}$ to [0, 1], formally defined by
\begin{equation}\label{eq_Massfunction1}
 m: 2^{U} \rightarrow [0, 1],
\end{equation}
which satisfies the following condition:
\begin{equation}\label{eq_Massfunction2}
 m(\emptyset) = 0\ and \sum\limits_{A \in 2^{U}} m(A) = 1.
\end{equation}
\end{definition}

In the Dempster--Shafer evidence theory, a mass function can be also called as a basic belief assignment (BBA).
If $m(A)$ is grater than 0, $A$ will be called as a focal element, and the union of all of the focal elements is called as the core of the mass function.

\begin{definition}(Belief function).

For a proposition $A \subseteq U$, the belief function $Bel: 2^{U} \rightarrow [0, 1]$ is defined as
\begin{equation}\label{eq_Belieffunction1}
 Bel(A) = \sum\limits_{B \subseteq A} m(B).
\end{equation}

The plausibility function $Pl: 2^{U} \rightarrow [0, 1]$  is defined as
\begin{equation}
 Pl(A) = 1 - Bel(\bar{A}) = \sum\limits_{B \cap A \neq \emptyset} m(B),
\end{equation}
where $\bar{A} = U - A$.
\end{definition}

Apparently, $Pl(A)$ is equal or greater than $Bel(A)$, where the function $Bel$ is the lower limit function of proposition $A$ and the function $Pl$ is the upper limit function of proposition $A$.

\begin{definition}(Dempster's rule of combination).

Let two BBAs $m_1$ and $m_2$ on the frame of discernment $U$ and assuming that these BBAs are independent, Dempster's rule of combination, denoted by $m = m_1 \oplus m_2$, which is called as the orthogonal sum, is defined as below:
\begin{equation}\label{eq_Dempsterrule1}
{
m(A) = \left\{ \begin{array}{l}
\begin{array}{*{20}{c}}
{\frac{1}{1-K} \sum\limits_{B \cap C = A} m_1(B) m_2(C),}&{{\kern 30pt} A \neq \emptyset,}
\end{array}\\
\begin{array}{*{20}{c}}
{0,}&{{\kern 136pt} A = \emptyset,}
\end{array}
\end{array} \right.
}\end{equation}
with
\begin{equation}\label{eq_Dempsterrule2}
{
K = \sum\limits_{B \cap C = \emptyset} m_1(B) m_2(C),
}\end{equation}
where $B$ and $C$ are also the elements of $2^{U}$, and $K$ is a constant that presents the conflict between two BBAs.
\end{definition}

Notice that, the Dempster's combination rule is only practicable for the two BBAs with the condition $K < 1$.

\subsection{Generalised Jensen-Shannon divergence measure}
Lin~\cite{lin1991divergence} introduced an information-theoretical based divergence measure among multi-probability distributions, called as Generalised Jensen-Shannon (GJS) divergence.
Most measures of difference are designed for two probability distributions.
Unlike others divergence measures, the main properties of GJS divergence are that, it does not require the condition of absolute continuity for the probability distributions involved;
and it is a measure of divergence for the overall difference of more than two distributions.
The main concepts are defined as below.

\begin{definition}(The GJS divergence among multi-probability distributions)~\cite{lin1991divergence,lamberti2008metric}.

Let $P$ be a set of probability distributions $\{P_1,P_2,\ldots,P_j,\ldots,P_k\}$ with a corresponding set of weights $\{\omega_1,\omega_2,\ldots,\omega_j,\ldots,\omega_k\}$, where $\sum^k_{j=1} \omega_j =1$ and $\omega_j \geq 0$.
The GJS divergence among the multi-probability distributions $\{P_1,\ldots,P_j,\ldots,P_k\}$ is denoted as:

\begin{equation}\label{eq_GJSdivergence1}
\begin{split}
GJS_\omega(P_1,\ldots,P_j,\ldots,P_k) = &H \left(\omega_1P_1 + \cdots + \omega_jP_j + \cdots  + \omega_kP_k\right) - \omega_1H(P_1) \\
&- \cdots  - \omega_jH(P_j) - \cdots - \omega_kH(P_k),
\end{split}
\end{equation}
with
\begin{equation}\label{eq_GJSdivergence2}
H(P_j)=-\sum_i p_{ji}\ log\ p_{ji},
\end{equation}
%and
%\begin{equation}\label{eq_GJSdivergence3}
%\sum_i p_{ji} = 1
%\end{equation}
where $\sum_i p_{ji} = 1$ $(i = 1, 2, \ldots, M; j = 1, 2, \ldots, k)$, and $H(P_j)$ is the Shannon entropy.

The $GJS_\omega(P_1,\ldots,P_j,\ldots,P_k)$ can be also expressed in the form
\begin{equation}\label{eq_GJSdivergence3}
\begin{split}
GJS_\omega(P_1,\ldots,P_j,\ldots,P_k) =
&\omega_1 \sum_i p_{1i}\ log \left(\frac{p_{1i}}{\omega_1p_{1i} + \omega_2 p_{2i} + \cdots + \omega_kp_{ki}}\right) + \\
&\cdots \\
&\omega_j \sum_i p_{ji}\ log \left(\frac{p_{ji}}{\omega_1p_{1i} + \omega_2 p_{2i} + \cdots + \omega_kp_{ki}}\right) + \\
&\cdots \\
&\omega_k \sum_i p_{ki}\ log \left(\frac{p_{ki}}{\omega_1p_{1i} + \omega_2 p_{2i} + \cdots + \omega_kp_{ki}}\right).
\end{split}
\end{equation}
\end{definition}

\newtheorem{myDef}{Remark}
\begin{myDef}
When $\omega_j=\frac{1}{k}$ $(1 \leq j \leq k)$, $GJS_{\omega_j=\frac{1}{k}}(P_1,\ldots,P_j,\ldots,P_k)$ is a special case of the GJS divergence, which means all of the multi-probability distributions $\{P_1,\ldots,P_j,\ldots,P_k\}$ have the same weights, denoted as
\begin{equation}\label{eq_GJSdivergence4}
\begin{split}
GJS_{\omega_j=\frac{1}{k}}(P_1,\ldots,P_j,\ldots,P_k) =
&\frac{1}{k} \sum_i p_{1i}\ log \left(\frac{p_{1i}}{\frac{1}{k}p_{1i} + \frac{1}{k}p_{2i} + \cdots + \frac{1}{k}p_{ki}}\right) + \\
&\cdots \\
&\frac{1}{k} \sum_i p_{ji}\ log \left(\frac{p_{ji}}{\frac{1}{k}p_{1i} + \frac{1}{k}p_{2i} + \cdots + \frac{1}{k}p_{ki}}\right) + \\
&\cdots \\
&\frac{1}{k} \sum_i p_{ki}\ log \left(\frac{p_{ki}}{\frac{1}{k}p_{1i} + \frac{1}{k}p_{2i} + \cdots + \frac{1}{k}p_{ki}}\right).
\end{split}
\end{equation}
\end{myDef}

There are some properties for the GJS divergence:

(1) $GJS_\omega(P_1,\ldots,P_j,\ldots,P_k)$ is zero if and only if all $P_j$ for which $\omega_j > 0$ are equal;

(2) $GJS_\omega(P_1,\ldots,P_j,\ldots,P_k)$ is symmetric and always well defined;

(3) its square root, $\sqrt{GJS_\omega(P_1,\ldots,P_j,\ldots,P_k)}$ verifies the triangle inequality.

(4) $GJS_\omega(P_1,\ldots,P_j,\ldots,P_k)$ is bounded, where $0 \leq GJS_\omega(P_1,\ldots,P_j,\ldots,P_k)$ $\leq \log_2 k$.

\section{Generalised belief divergence measure}\label{BJSdivergence}
In Dempster--Shafer evidence theory, it is still an open issue about how to measure the discrepancy and conflict among evidences, which is significant for the fusion of multiple evidences.
Dempster--Shafer evidence theory, as a generalization of probability theory, allocates the probability into the subsets of the set including multiple objects, not just to an individual object.
In order to measure the divergence for multi-BBAs in Dempster--Shafer evidence theory, a novel generalised divergence measure called Generalised Belief Jensen-Shannon (GBJS) divergence is proposed for the belief function based on the Generalised Jensen-Shannon divergence.
The basic concepts are defined as below.

\begin{definition}(The GBJS divergence among multi-BBAs).

Let $A_i$ be a hypothesis of the belief function $m$, and let $\{m_1,\ldots,m_j,\ldots,m_k\}$ be a set of BBAs with a corresponding set of reliability weights $\{\omega_1,\ldots,\omega_j,\ldots,\omega_k\}$ on the same frame of discernment $\Omega$, containing $N$ mutually exclusive and exhaustive hypotheses.
The GBJS divergence among the $k$ BBAs $\{m_1,\ldots,m_j,\ldots,m_k\}$ is denoted as:

\begin{equation}\label{eq_GBJSdivergence1}
\begin{split}
GBJS_\omega(m_1,\ldots,m_j,\ldots,m_k) = &H \left(\omega_1m_1 + \cdots + \omega_jm_j + \cdots  + \omega_km_k\right) - \omega_1H(m_1) \\
&- \cdots  - \omega_jH(m_j) - \cdots - \omega_kH(m_k),
\end{split}
\end{equation}
with
\begin{equation}\label{eq_GBJSdivergence2}
H(m_j)=-\sum_i m_j(A_i)\ log\ m_j(A_i),
\end{equation}
where $\sum_i m_j(A_i) = 1$ $(i = 1, 2, \ldots, M; j = 1, 2, \ldots, k)$, and $H(m_j)$ is the Shannon entropy.

$GBJS_\omega(m_1,\ldots,m_j,\ldots,m_k)$ can be also expressed in the following form
\begin{equation}\label{eq_GBJSdivergence3}
\begin{split}
GBJS_\omega (m_1,&\ldots,m_j,\ldots,m_k) = \\
&\omega_1 \sum_i m_1(A_i)\ log \left(\frac{m_1(A_i)}{\omega_1m_1(A_i) + \omega_2 m_2(A_i) + \cdots + \omega_km_k(A_i)}\right) + \\
&\cdots \\
&\omega_j \sum_i m_j(A_i)\ log \left(\frac{m_j(A_i)}{\omega_1m_1(A_i) + \omega_2 m_2(A_i) + \cdots + \omega_km_k(A_i)}\right) + \\
&\cdots \\
&\omega_k \sum_i m_k(A_i)\ log \left(\frac{m_k(A_i)}{\omega_1m_1(A_i) + \omega_2 m_2(A_i) + \cdots + \omega_km_k(A_i)}\right).
\end{split}
\end{equation}
\end{definition}

\begin{myDef}
When $\omega_j=\frac{1}{k}$ $(1 \leq j \leq k)$, $GBJS_{\omega_j=\frac{1}{k}}(m_1,\ldots,m_j,\ldots,m_k)$ is a special case of the GBJS divergence, which indicates that all of the multi-BBAs $\{m_1,\ldots,m_j,\ldots,m_k\}$ have the same reliability weights, denoted as
\begin{equation}\label{eq_GBJSdivergence4}
\begin{split}
GBJS_{\omega_j=\frac{1}{k}}(m_1,&\ldots,m_j,\ldots,m_k) = \\
&\frac{1}{k} \sum_i m_1(A_i)\ log \left(\frac{m_1(A_i)}{\frac{1}{k}m_1(A_i) + \frac{1}{k}m_2(A_i) + \cdots + \frac{1}{k}m_k(A_i)}\right) + \\
&\cdots \\
&\frac{1}{k} \sum_i m_j(A_i)\ log \left(\frac{m_j(A_i)}{\frac{1}{k}m_1(A_i) + \frac{1}{k}m_2(A_i) + \cdots + \frac{1}{k}m_k(A_i)}\right) + \\
&\cdots \\
&\frac{1}{k} \sum_i m_k(A_i)\ log \left(\frac{m_k(A_i)}{\frac{1}{k}m_1(A_i) + \frac{1}{k}m_2(A_i) + \cdots + \frac{1}{k}m_k(A_i)}\right).
\end{split}
\end{equation}
\end{myDef}

Note that when a BBA is allocated with zero, the fraction value is going to infinity and the value of its logarithm is also close to infinity.
In such a case, the proposed method cannot work, therefore, a very small value $1 \times 10^{-12}$ is allocated to the BBA instead of zero value, which has been proven that it did not affect the results~\cite{IOPORT.06556237}.

The Generalised Belief Jensen-Shannon divergence is similar with the Generalised Jensen-Shannon divergence in form.
Whereas, the Generalised Belief Jensen-Shannon divergence is designed for the mass function, rather than for the probability distribution function.
In such a situation where all of the hypothesis of belief functions are allocated to single elements, the BBAs turn into probabilities, so that the Generalised Belief Jensen-Shannon divergence degenerates to the Generalised Jensen-Shannon divergence.

The properties of the GBJS divergence can be inferred as below:

(1) $GBJS_\omega(m_1,\ldots,m_j,\ldots,m_k)$ is zero if and only if all $m_j$ for which $\omega_j > 0$ are equal;

(2) $GBJS_\omega(m_1,\ldots,m_j,\ldots,m_k)$ is symmetric and always well defined;

(3) its square root, $\sqrt{GBJS_\omega(m_1,\ldots,m_j,\ldots,m_k)}$ verifies the triangle inequality.

(4) $GBJS_\omega(m_1,\ldots,m_j,\ldots,m_k)$ is bounded, where $0 \leq GBJS_\omega(m_1,\ldots,m_j,$ $\ldots,m_k) \leq \log_2 k$.

\newtheorem{exmp}{Example}

\begin{exmp}\label{exa_BJS1}
\rm Supposing that there are three BBAs $m_1$, $m_2$ and $m_3$ with the same reliability weights, namely, $\omega_1=\omega_2=\omega_3=\frac{1}{3}$ in the frame of discernment $\Omega=\{A,B,C\}$ which is complete, and the three BBAs are given as follows:
\end{exmp}

\begin{tabular}[t]{l}
$m_1:$ $m_1(A)=0.6$, $m_1(B)=0.1$, $m_1(C)=0.3$;\\
$m_2:$ $m_2(A)=0.6$, $m_2(B)=0.1$, $m_2(C)=0.3$;\\
$m_3:$ $m_3(A)=0.6$, $m_3(B)=0.1$, $m_3(C)=0.3$.\\
\specialrule{0em}{6pt}{6pt}
\end{tabular}

As shown in Example~\ref{exa_BJS1}, it can be see that $m_1$, $m_2$ and $m_3$ have the same BBAs with each other, where $m_1(A)=m_2(A)=m_3(A)=0.6$, $m_1(B)=m_2(B)=m_3(B)=0.1$, and $m_1(C)=m_2(C)=m_3(C)=0.3$.
Then, the specific calculation processes of the Generalised Belief Jensen-Shannon divergence $GBJS(m_1,m_2,m_3)$ are listed as follows:

\begin{align*}
GBJS_\omega(m_1,m_2,m_3) =
&\frac{1}{3}\times0.6\times \log(\frac{0.6}{\frac{1}{3}\times0.6+\frac{1}{3}\times0.6+\frac{1}{3}\times0.6})+\\
&\frac{1}{3}\times0.6\times \log(\frac{0.6}{\frac{1}{3}\times0.6+\frac{1}{3}\times0.6+\frac{1}{3}\times0.6})+\\
&\frac{1}{3}\times0.6\times \log(\frac{0.6}{\frac{1}{3}\times0.6+\frac{1}{3}\times0.6+\frac{1}{3}\times0.6})+\\
&\frac{1}{3}\times0.1\times \log(\frac{0.1}{\frac{1}{3}\times0.1+\frac{1}{3}\times0.1+\frac{1}{3}\times0.1})+\\
&\frac{1}{3}\times0.1\times \log(\frac{0.1}{\frac{1}{3}\times0.1+\frac{1}{3}\times0.1+\frac{1}{3}\times0.1})+\\
&\frac{1}{3}\times0.1\times \log(\frac{0.1}{\frac{1}{3}\times0.1+\frac{1}{3}\times0.1+\frac{1}{3}\times0.1})+\\
&\frac{1}{3}\times0.3\times \log(\frac{0.3}{\frac{1}{3}\times0.3+\frac{1}{3}\times0.3+\frac{1}{3}\times0.3})+\\
&\frac{1}{3}\times0.3\times \log(\frac{0.3}{\frac{1}{3}\times0.3+\frac{1}{3}\times0.3+\frac{1}{3}\times0.3})+\\
&\frac{1}{3}\times0.3\times \log(\frac{0.3}{\frac{1}{3}\times0.3+\frac{1}{3}\times0.3+\frac{1}{3}\times0.3})\\
= &0.
\end{align*}

This example verifies that when $m_1$, $m_2$ and $m_3$ have the same BBAs with each other, the Generalised Belief Jensen-Shannon divergence among the BBAs $m_1$, $m_2$ and $m_3$ is 0 which accords with an intuitionistic result.

\begin{exmp}\label{exa_BJS2}
\rm Supposing that there are three BBAs $m_1$, $m_2$ and $m_3$ with a reliability weight, $\omega_1=0.5$, $\omega_2=0.4$, and $\omega_3=0.1$, respectively, in the frame of discernment $\Omega=\{A,B,C\}$ which is complete, and the three BBAs are given as follows:
\end{exmp}

\begin{tabular}[t]{l}
$m_1:$ $m_1(A)=0.7$, $m_1(B)=0.1$, $m_1(C)=0.2$;\\
$m_2:$ $m_2(A)=0.6$, $m_2(B)=0.1$, $m_2(C)=0.3$;\\
$m_3:$ $m_3(A)=0.3$, $m_3(B)=0.2$, $m_3(C)=0.5$.\\
\specialrule{0em}{6pt}{6pt}
\end{tabular}

As shown in Example~\ref{exa_BJS2}, we can notice that $m_1$ and $m_2$ have relatively large reliability weights, namely, $\omega_1=0.5$ and $\omega_2=0.4$;
and they have relatively large belief values to support the object $A$, where $m_1(A)=0.7$ and $m_2(A)=0.6$, while $m_3$ with a relatively low reliability weight $\omega_3=0.1$, has $m_3(C)=0.5$ belief value to support the object $C$.
The Generalised Belief Jensen-Shannon divergence $GBJS(m_1,m_2,m_3)$ for $m_1$, $m_2$ and $m_3$ is calculated as follows:
\begin{align*}
GBJS_\omega(m_1,m_2,m_3) =
&0.5\times0.7\times \log(\frac{0.7}{0.5\times0.7+0.4\times0.6+0.1\times0.3})+\\
&0.5\times0.1\times \log(\frac{0.1}{0.5\times0.1+0.4\times0.1+0.1\times0.2})+\\
&0.5\times0.2\times \log(\frac{0.2}{0.5\times0.2+0.4\times0.3+0.1\times0.5})+\\
&0.4\times0.6\times \log(\frac{0.6}{0.5\times0.7+0.4\times0.6+0.1\times0.3})+\\
&0.4\times0.1\times \log(\frac{0.1}{0.5\times0.1+0.4\times0.1+0.1\times0.2})+\\
&0.4\times0.3\times \log(\frac{0.3}{0.5\times0.2+0.4\times0.3+0.1\times0.5})+\\
&0.1\times0.3\times \log(\frac{0.3}{0.5\times0.7+0.4\times0.6+0.1\times0.3})+\\
&0.1\times0.2\times \log(\frac{0.2}{0.5\times0.1+0.4\times0.1+0.1\times0.2})+\\
&0.1\times0.5\times \log(\frac{0.5}{0.5\times0.2+0.4\times0.3+0.1\times0.5})\\
= &0.0428.
\end{align*}

On the other hand, based on Eq.~(\ref{eq_GBJSdivergence3}), the Generalised Belief Jensen-Shannon divergence $GBJS(m_1,m_3,m_2)$, $GBJS(m_2,m_1,m_3)$, $GBJS(m_2,m_3,m_1)$, $GBJS(m_3,m_1,m_2)$ and $GBJS(m_3,m_2,m_1)$ can also be obtained:

\begin{align*}
GBJS_\omega(m_1,m_2,m_3)
= &GBJS(m_1,m_3,m_2) \\
= &GBJS(m_2,m_1,m_3) \\
= &GBJS(m_2,m_3,m_1) \\
= &GBJS(m_3,m_1,m_2) \\
= &GBJS(m_3,m_2,m_1) \\
= &0.0428.
\end{align*}

From the above results, it can be concluded that the symmetric property of the Generalised Belief Jensen-Shannon divergence measure method is verified in this example.

\begin{exmp}\label{exa_BJS3}
\rm Supposing that there are four BBAs $m_1$, $m_2$, $m_3$ and $m_4$ with a reliability weight, $\omega_1=\omega_2=\omega_3=\omega_4=\frac{1}{4}$, respectively, in the frame of discernment $\Omega=\{A,B\}$ which is complete, and the four BBAs are given as follows:
\end{exmp}

\begin{tabular}[t]{l}
$m_1:$ $m_1(A)=0.98$, $m_1(B)=0.01$, $m_1(A,B)=0.01$;\\
$m_2:$ $m_2(A)=0.97$, $m_2(B)=0.02$, $m_2(A,B)=0.01$;\\
$m_3:$ $m_3(A)=0.90$, $m_3(B)=0.05$, $m_3(A,B)=0.05$;\\
$m_4:$ $m_4(A)=0.01$, $m_4(B)=0.98$, $m_4(A,B)=0.01$.\\
\specialrule{0em}{6pt}{6pt}
\end{tabular}

As shown in Example~\ref{exa_BJS3}, we can see that $m_1$, $m_2$ and $m_3$ have great belief values to support the object $A$, where $m_1(A)=0.98$, $m_2(A)=0.97$ and $m_3(A)=0.90$.
On the contrary, $m_4$ has a great belief value to support the object $B$, where $m_4(B)=0.98$.
For the Generalised Belief Jensen-Shannon divergence $GBJS(m_1,m_2,m_3)$ among the BBAs $m_1$, $m_2$ and $m_3$, their corresponding reliability weights can be normalised as $\omega_1=\omega_2=\omega_3=\frac{1}{3}$;
then, the $GBJS(m_1,m_2,m_3)$ is calculated below:

\begin{align*}
GBJS(m_1,m_2,m_3) =
&\frac{1}{3}\times0.98\times\log(\frac{0.98}{\frac{1}{3}\times0.98+\frac{1}{3}\times0.97+\frac{1}{3}\times0.90})+\\
&\frac{1}{3}\times0.01\times\log(\frac{0.01}{\frac{1}{3}\times0.01+\frac{1}{3}\times0.02+\frac{1}{3}\times0.05})+\\
&\frac{1}{3}\times0.01\times\log(\frac{0.01}{\frac{1}{3}\times0.01+\frac{1}{3}\times0.01+\frac{1}{3}\times0.05})+\\
&\frac{1}{3}\times0.97\times\log(\frac{0.97}{\frac{1}{3}\times0.98+\frac{1}{3}\times0.97+\frac{1}{3}\times0.90})+\\
&\frac{1}{3}\times0.02\times\log(\frac{0.02}{\frac{1}{3}\times0.01+\frac{1}{3}\times0.02+\frac{1}{3}\times0.05})+\\
&\frac{1}{3}\times0.01\times\log(\frac{0.01}{\frac{1}{3}\times0.01+\frac{1}{3}\times0.01+\frac{1}{3}\times0.05})+\\
&\frac{1}{3}\times0.90\times\log(\frac{0.90}{\frac{1}{3}\times0.98+\frac{1}{3}\times0.97+\frac{1}{3}\times0.90})+\\
&\frac{1}{3}\times0.05\times\log(\frac{0.05}{\frac{1}{3}\times0.01+\frac{1}{3}\times0.02+\frac{1}{3}\times0.05})+\\
&\frac{1}{3}\times0.05\times\log(\frac{0.05}{\frac{1}{3}\times0.01+\frac{1}{3}\times0.01+\frac{1}{3}\times0.05})\\
= &0.0188.
\end{align*}

On the other hand,the Generalised Belief Jensen-Shannon divergence $GBJS(m_1,m_2,m_4)$ among the BBAs $m_1$, $m_2$ and $m_4$ is computed as follows:

\begin{align*}
GBJS(m_1,m_2,m_4) =
&\frac{1}{3}\times0.98\times\log(\frac{0.98}{\frac{1}{3}\times0.98+\frac{1}{3}\times0.97+\frac{1}{3}\times0.01})+\\
&\frac{1}{3}\times0.01\times\log(\frac{0.01}{\frac{1}{3}\times0.01+\frac{1}{3}\times0.02+\frac{1}{3}\times0.98})+\\
&\frac{1}{3}\times0.01\times\log(\frac{0.01}{\frac{1}{3}\times0.01+\frac{1}{3}\times0.01+\frac{1}{3}\times0.01})+\\
&\frac{1}{3}\times0.97\times\log(\frac{0.97}{\frac{1}{3}\times0.98+\frac{1}{3}\times0.97+\frac{1}{3}\times0.01})+\\
&\frac{1}{3}\times0.02\times\log(\frac{0.02}{\frac{1}{3}\times0.01+\frac{1}{3}\times0.02+\frac{1}{3}\times0.98})+\\
&\frac{1}{3}\times0.01\times\log(\frac{0.01}{\frac{1}{3}\times0.01+\frac{1}{3}\times0.01+\frac{1}{3}\times0.01})+\\
&\frac{1}{3}\times0.01\times\log(\frac{0.01}{\frac{1}{3}\times0.98+\frac{1}{3}\times0.97+\frac{1}{3}\times0.01})+\\
&\frac{1}{3}\times0.98\times\log(\frac{0.98}{\frac{1}{3}\times0.01+\frac{1}{3}\times0.02+\frac{1}{3}\times0.98})+\\
&\frac{1}{3}\times0.01\times\log(\frac{0.01}{\frac{1}{3}\times0.01+\frac{1}{3}\times0.01+\frac{1}{3}\times0.01})\\
= &0.8148.
\end{align*}

Moreover, the Generalised Belief Jensen-Shannon divergence $GBJS(m_2,m_3,m_4)$ among the BBAs $m_2$, $m_3$ and $m_4$ is computed as follows:

\begin{align*}
GBJS(m_2,m_3,m_4) =
&\frac{1}{3}\times0.97\times\log(\frac{0.97}{\frac{1}{3}\times0.97+\frac{1}{3}\times0.90+\frac{1}{3}\times0.01})+\\
&\frac{1}{3}\times0.02\times\log(\frac{0.02}{\frac{1}{3}\times0.02+\frac{1}{3}\times0.05+\frac{1}{3}\times0.98})+\\
&\frac{1}{3}\times0.01\times\log(\frac{0.01}{\frac{1}{3}\times0.01+\frac{1}{3}\times0.05+\frac{1}{3}\times0.01})+\\
&\frac{1}{3}\times0.90\times\log(\frac{0.90}{\frac{1}{3}\times0.97+\frac{1}{3}\times0.90+\frac{1}{3}\times0.01})+\\
&\frac{1}{3}\times0.05\times\log(\frac{0.05}{\frac{1}{3}\times0.02+\frac{1}{3}\times0.05+\frac{1}{3}\times0.98})+\\
&\frac{1}{3}\times0.05\times\log(\frac{0.05}{\frac{1}{3}\times0.01+\frac{1}{3}\times0.05+\frac{1}{3}\times0.01})+\\
&\frac{1}{3}\times0.01\times\log(\frac{0.01}{\frac{1}{3}\times0.97+\frac{1}{3}\times0.01+\frac{1}{3}\times0.01})+\\
&\frac{1}{3}\times0.98\times\log(\frac{0.98}{\frac{1}{3}\times0.02+\frac{1}{3}\times0.98+\frac{1}{3}\times0.98})+\\
&\frac{1}{3}\times0.01\times\log(\frac{0.01}{\frac{1}{3}\times0.01+\frac{1}{3}\times0.01+\frac{1}{3}\times0.01})\\
= &0.7617.
\end{align*}

After that, their corresponding square root values can be calculated as follows:

\begin{tabular}[t]{l}
$\sqrt{GBJS(m_1,m_2,m_3)}=\sqrt{0.0188}=0.1370$;\\
$\sqrt{GBJS(m_1,m_2,m_4)}=\sqrt{0.8148}=0.9027$;\\
$\sqrt{GBJS(m_2,m_3,m_4)}=\sqrt{0.7617}=0.8727$.\\
\specialrule{0em}{6pt}{6pt}
\end{tabular}

It can be obtained that $\sqrt{GBJS(m_1,m_2,m_3)}+\sqrt{GBJS(m_2,m_3,m_4)}=1.0098$, so that $\sqrt{GBJS(m_1,m_2,m_4)}<\sqrt{GBJS(m_1,m_2,m_3)}+\sqrt{GBJS(m_2,m_3,m_4)}$, which satisfies the triangle inequality property of the Generalised Belief Jensen-Shannon divergence measure method.

\section{The proposed method}\label{Proposed method}
\begin{figure}[!htpb]
\centering
\includegraphics[width=12cm,clip]{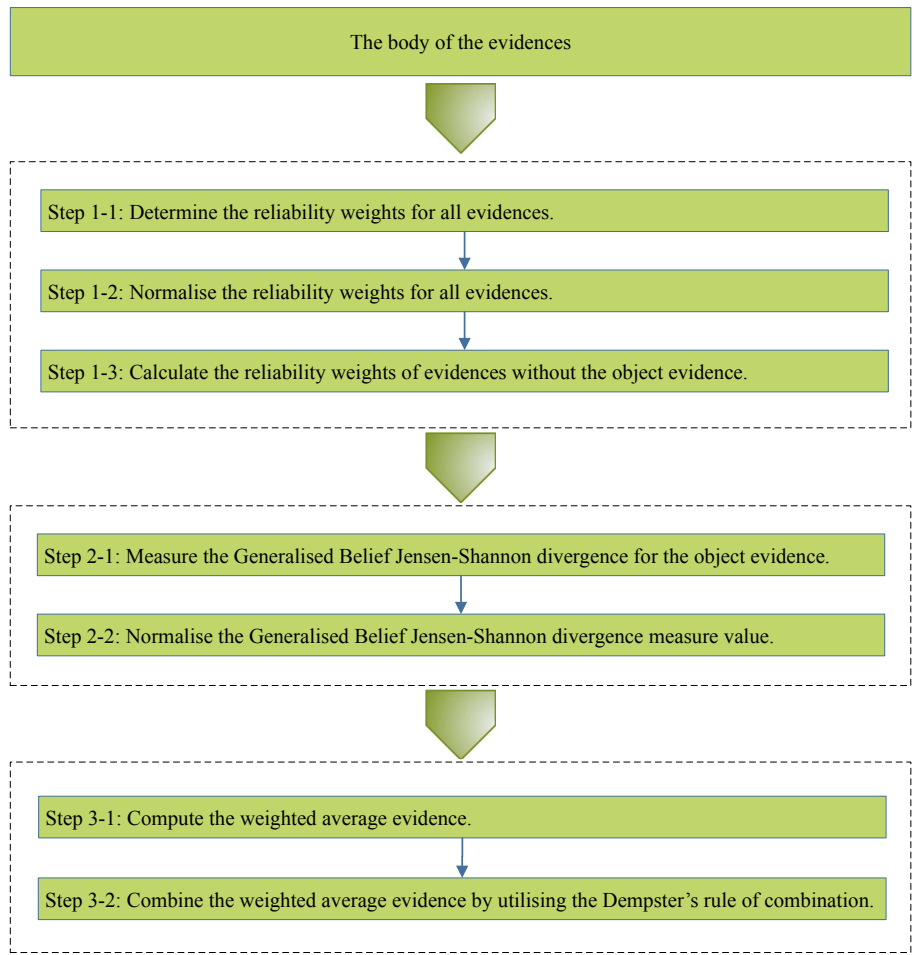}
 \caption{The flowchart of the proposed method.}\label{flowchart}
\end{figure}

In this paper, a novel multi-sensor data fusion method is proposed on the basis of the Generalised Belief Jensen-Shannon divergence measure of evidences.
It consists of the following three parts.
Firstly, the reliability weights of evidences are determined by considering the sufficiency index and importance index of the evidences.
After that, on account of the reliability weights of evidences, the presented Generalised Belief Jensen-Shannon (GBJS) divergence is utilised to measure the support degrees of evidences by taking into account the discrepancy and conflict degree among multiple evidences, where the support degree is considered as the final weight.
Ultimately, the final weights of the evidences are used to adjust the bodies of the evidences before using the Dempster's combination rule.
The flowchart of the proposed method is shown in Fig.~\ref{flowchart}.

\subsection{Calculate the reliability weights of evidences}
The sensor reliability plays an important role to comprehend and quantify the performance of the sensor, which is used to judge whether the fusion result is reasonable or not, such as work efficiency, accuracy and experts with different knowledge.
The sensor reliability may be influenced by the technical factors and noise, like manufacturing craft, material, principle, and so on.
Generally, the sensor reliability is assessed by comparing the output value with the actual value of the sensor via long term practical applications.
In this paper, the sufficiency and importance of evidence in~\cite{fan2006fault} are adopted to measure the reliability of sensors.
Note that if the sufficiency and importance of evidence are not given in some certain cases, the reliability weights of evidences are supposed to be averaged.

\begin{enumerate}[{Step 1-}1:]
\item
The reliability weight of evidence $m_j$ $(j = 1, 2, \ldots, k$), denoted as $\omega_j$ is determined by the following formula:
\begin{equation}\label{eq_reliability}
{
\omega_j = \mu_j \times v_j, \quad 1 \leq j \leq k,
}\end{equation}
where $\mu_j$ represents the sufficiency of evidence $m_j$, and $v_j$ denotes the importance of evidence $m_j$.

When the sufficiency and importance of evidence are not given in some certain cases, the reliability weights of $k$ evidences is calculated by:
\begin{equation}\label{eq_normalisedreliability}
{
\omega_j = \frac{1}{k}, \quad  1 \leq j \leq k.
}\end{equation}

\item
The reliability weight of evidence $m_j$ is normalised below, denoted as $\overline{\omega}_j$:
\begin{equation}\label{eq_normalisedreliabilityweight1}
{
\overline{\omega}_j = \frac{\omega_j}{\sum_{s=1}^k \omega_s}, \quad  1 \leq j \leq k.
}\end{equation}

\item
To measure the GBJS divergence of specific evidences for an object evidence $m_j$, the reliability weight of evidence $m_i$ $(i = 1, 2, \ldots, k; i \neq j)$, denoted as $\widetilde{\omega}_i$ is calculated by:
\begin{equation}\label{eq_normalisedreliabilityweight2}
{
\widetilde{\omega}_i = \frac{\omega_i}{\sum_{s=1; s \neq j}^k \omega_s}, \quad  1 \leq i \leq k; i \neq j.
}\end{equation}
\end{enumerate}

\subsection{Measure the support degrees of evidences}
In this section, the support degrees of evidences are measured based on the Generalised Belief Jensen-Shannon (GBJS) divergence of evidences.
The question is that how can we leverage the GBJS divergence of evidences to measure the support degrees of evidences?
To achieve this goal, the following steps are considered.

\begin{enumerate}[{Step 2-}1:]
\item
The support degree of the evidence $m_j$ $(j = 1, 2, \ldots, k$), denoted as $Sup(m_j)$ is measured by:
\begin{equation}\label{eq_supportdegree1}
\begin{aligned}
Sup(m_j) = &GBJS_{\widetilde{\omega}}(m_1,\ldots,m_{j-1},m_{j+1},\ldots,m_k),
\end{aligned}
\end{equation}
because the Generalised Belief Jensen-Shannon divergence measure $GBJS_\omega(m_1,\ldots,m_{j-1},m_{j+1},\ldots,m_k)$ without $m_j$ implies how much difference between the evidence $m_j$ and other evidences.
Specifically, if the evidence $m_j$ highly conflicts with other evidences, the support degree $Sup(m_j)$ excepting itself is supposed to be small, since the divergence measure value is relatively small under the situation where the highly conflicting evidence is excluded.
In a word, the more conflict the evidence $m_j$ has, the less support degree $Sup(m_j)$ has.

In the case where the sufficiency and importance of evidence are not provided, the support degree of the evidence $m_j$ $(j = 1, 2, \ldots, k$), denoted as $Sup(m_j)$ is measured by:
\begin{equation}\label{eq_supportdegree2}
\begin{aligned}
Sup(m_j) = &GBJS_{{\widetilde{\omega}}=\frac{1}{k-1}}(m_1,\ldots,m_{j-1},m_{j+1},\ldots,m_k).
\end{aligned}
\end{equation}

\item
The support degree $Sup(m_j)$ of evidence $m_j$ is normalised below, denoted as $\widetilde{S}up_j$, which is considered as the final weight:
\begin{equation}\label{eq_normalisedsupportdegree}
{
\widetilde{S}up_j = \frac{Sup(m_j)}{\sum_{s=1}^{k} Sup(m_s)}, \quad \quad  1 \leq j \leq k.
}\end{equation}
\end{enumerate}

\subsection{Fuse the evidences based on the Dempster's combination rule}

\begin{enumerate}[{Step 3-}1:]
\item
In view of the final weight $\widetilde{S}up_j$ of the evidence $m_j$, the weighted average evidence, denoted as $WAE(m)$ can be generated as follows:
\begin{equation}\label{eq_WAE}
{
WAE(m) = \sum\limits_{i=1}^{k} (\widetilde{S}up_j \times m_j), \quad 1 \leq j \leq k.
}\end{equation}

\item
The weighted average evidence $WAE(m)$ is fused via the Dempster's combination rule of Eq.~(\ref{eq_Dempsterrule1}) by $k - 1$ times, when $k$ number of evidences exist.
Ultimately, the final combination result of multiple evidences can be produced.
\end{enumerate}

\section{Application}\label{Application}
In this section, the proposed method is applied to a motor rotor fault diagnosis, in which the practical data from~\cite{jiang2016sensordiagnosis} are adopted for the comparison of the related method.

\subsection{Problem Statement}
Supposing that the frame of discernment $\Theta$ = $\{Rotor\ unbalance$, $Rotor\ misalignment$, $Pedestal\ looseness\}$ includes three fault types $\{F_1, F_2, F_3\}$ for a motor rotor.
The set of vibration acceleration sensors $S$ = $\{S_1, S_2, S_3\}$ is placed at different places to gather the vibration signals.
The acceleration vibration frequency amplitudes at $1X$, $2X$ and $3X$ frequencies are regarded as the fault feature variables.
The collected sensor reports at $1X$, $2X$ and $3X$ frequencies modeled as BPAs are provided in Tables~\ref{app123X}, in which $m_1(\cdot)$, $m_2(\cdot)$ and $m_3(\cdot)$ denote the BPAs from the three vibration acceleration sensors $S_1$, $S_2$ and $S_3$.

\begin{sidewaystable}[!htbp]\renewcommand{\arraystretch}{1.2}
%\begin{table}[!h]\renewcommand{\arraystretch}{1.2}
{\footnotesize
\caption{The gathered sensor reports at $1X$, $2X$ and $3X$ frequencies modelled as BPAs.}\label{app123X}
\begin{tabular*}{\columnwidth}{@{\extracolsep{\fill}}@{~~}lllllllllll@{~~}}
\toprule
Frequencies  & \multicolumn{4}{c}{$1X$} & \multicolumn{2}{c}{$2X$} & \multicolumn{4}{c}{$3X$} \\ \midrule
BBAs         &$\{F_2\}$&$\{F_3\}$&$\{F_1,F_2\}$&$\{F_1,F_2,F_3\}$&$\{F_2\}$&$\{F_1,F_2,F_3\}$&$\{F_1\}$&$\{F_2\}$&$\{F_1,F_2\}$&$\{F_1,F_2,F_3\}$ \\\midrule
$m_1(\cdot)$ & 0.8176 & 0.0003 & 0.1553 & 0.0268& 0.6229 & 0.3771& 0.3666 & 0.4563 & 0.1185 & 0.0586 \\
$m_2(\cdot)$ & 0.5658 & 0.0009 & 0.0646 & 0.3687& 0.7660 & 0.2341& 0.2793 & 0.4151 & 0.2652 & 0.0404 \\
$m_3(\cdot)$ & 0.2403 & 0.0004 & 0.0141 & 0.7452& 0.8598 & 0.1402& 0.2897 & 0.4331 & 0.2470 & 0.0302 \\
\bottomrule
\end{tabular*}
}
%\end{table}
\end{sidewaystable}

\subsection{Implementation of the proposed method}
By implementing the proposed method, the weighted average evidences in terms of motor rotor fault diagnosis at $1X$, $2X$ and $3X$ frequencies can be obtained, respectively, as shown in Tables~\ref{weightedaverageevidence}.

\begin{sidewaystable}[!htbp]\renewcommand{\arraystretch}{1.2}
%\begin{table}[!h]\renewcommand{\arraystretch}{1.2}
{\footnotesize
\caption{The weighted average evidences at $1X$, $2X$ and $3X$ frequencies.}\label{weightedaverageevidence}
\begin{tabular*}{\columnwidth}{@{\extracolsep{\fill}}@{~~}lllllllllll@{~~}}
\toprule
Frequencies& \multicolumn{4}{c}{$1X$} & \multicolumn{2}{c}{$2X$} & \multicolumn{4}{c}{$3X$} \\ \midrule
BBAs     &$\{F_2\}$&$\{F_3\}$&$\{F_1,F_2\}$&$\{F_1,F_2,F_3\}$&$\{F_2\}$&$\{F_1,F_2,F_3\}$&$\{F_1\}$&$\{F_2\}$&$\{F_1,F_2\}$&$\{F_1,F_2,F_3\}$ \\\midrule
$WAE(m)$&0.5332&0.0007&0.0671&0.3990&0.7677&0.2324&0.2864&0.4253&0.2529&0.0354\\
\bottomrule
\end{tabular*}
}
%\end{table}
\end{sidewaystable}

After that, the weighted average evidences in terms of motor rotor fault diagnosis at $1X$, $2X$ and $3X$ frequencies are fused, respectively, via leveraging the Dempster's rule of combination by two times.

Finally, the combination results for motor rotor fault diagnosis at $1X$, $2X$ and $3X$ frequencies are generated, respectively, as shown in Tables~\ref{Fusionresults}.

\begin{sidewaystable}[!htbp]\renewcommand{\arraystretch}{1.2}
%\begin{table}[!h]\renewcommand{\arraystretch}{1.2}
{\footnotesize
\caption{Fusion results of different methods for motor rotor fault diagnosis.}\label{Fusionresults}
\begin{tabular*}{\columnwidth}{@{\extracolsep{\fill}}@{~~}llllllllllll@{~~}}
\toprule
Frequencies& \multicolumn{4}{c}{$1X$} & \multicolumn{2}{c}{$2X$} & \multicolumn{4}{c}{$3X$}&\multirow{2}{*}{Target} \\ \cmidrule(l){1-11}
Methods    &$\{F_2\}$&$\{F_3\}$&$\{F_1,F_2\}$&$\{F_1,F_2,F_3\}$&$\{F_2\}$&$\{F_1,F_2,F_3\}$&$\{F_1\}$&$\{F_2\}$&$\{F_1,F_2\}$&$\{F_1,F_2,F_3\}$ \\\midrule
Jiang et al.~\cite{jiang2016sensordiagnosis}
&0.8861&0.0002&0.0582&0.0555&0.9621&0.0371&0.3384&0.5904&0.0651&0.0061&$F_2$ \\
Proposed method
&0.8982&0.0003&0.0378&0.0636&0.9877&0.0126&0.3266&0.6365&0.0368&0.0001&$F_2$\\
\bottomrule
\end{tabular*}
}
%\end{table}
\end{sidewaystable}

\begin{figure}
\centering
\subfigure[]{
\label{exp4:a} %% label for first subfigure
\includegraphics[width=.475\linewidth,clip]{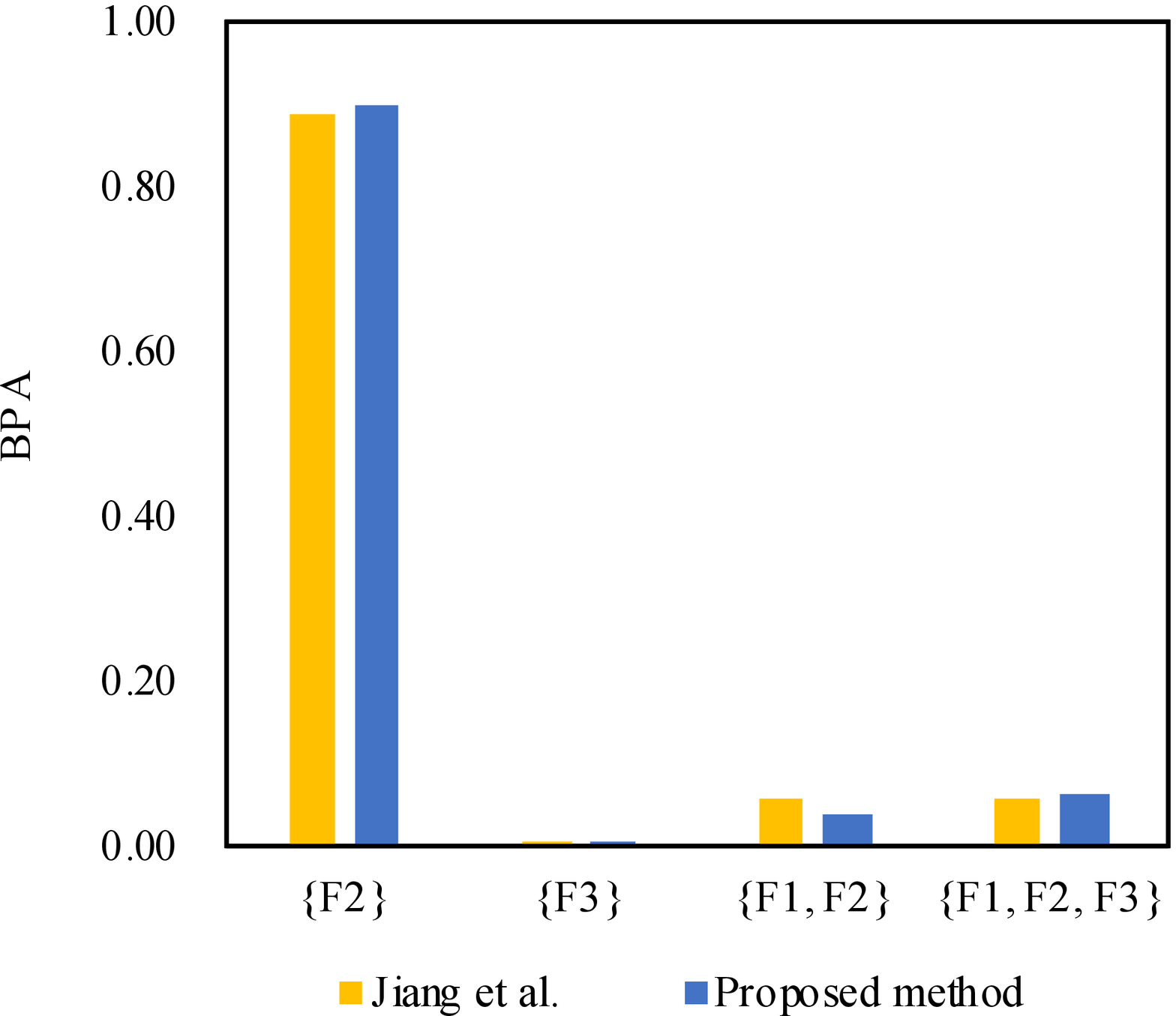}
}
\subfigure[]{
\label{exp4:b} %% label for second subfigure
\centering
\includegraphics[width=.475\linewidth,clip]{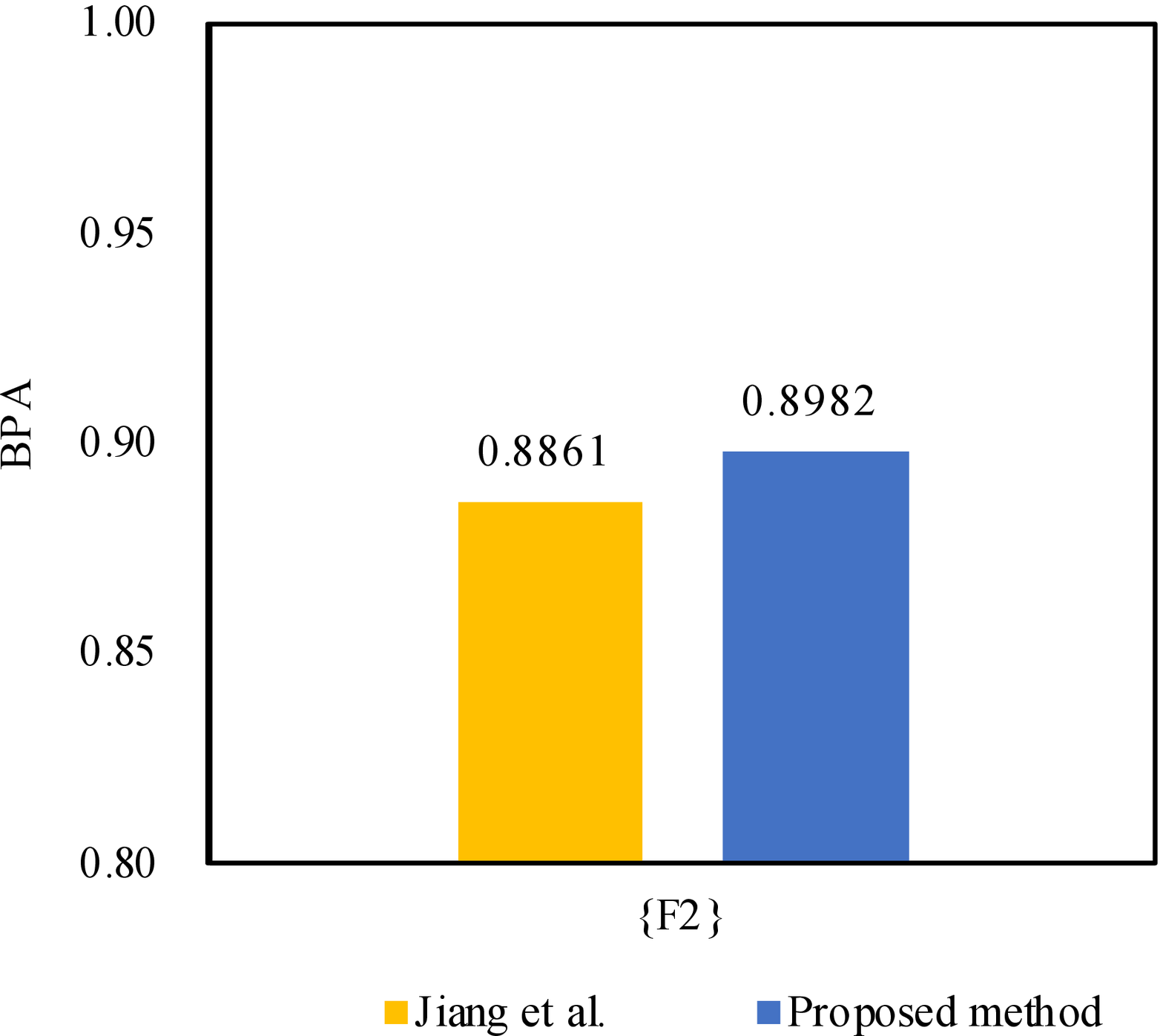}
}
\caption{The comparison of different methods for motor rotor fault diagnosis at $1X$ frequency.}
\label{1X} %% label for entire figure
\end{figure}

\begin{figure}
\centering
\subfigure[]{
\label{exp4:a} %% label for first subfigure
\includegraphics[width=.475\linewidth,clip]{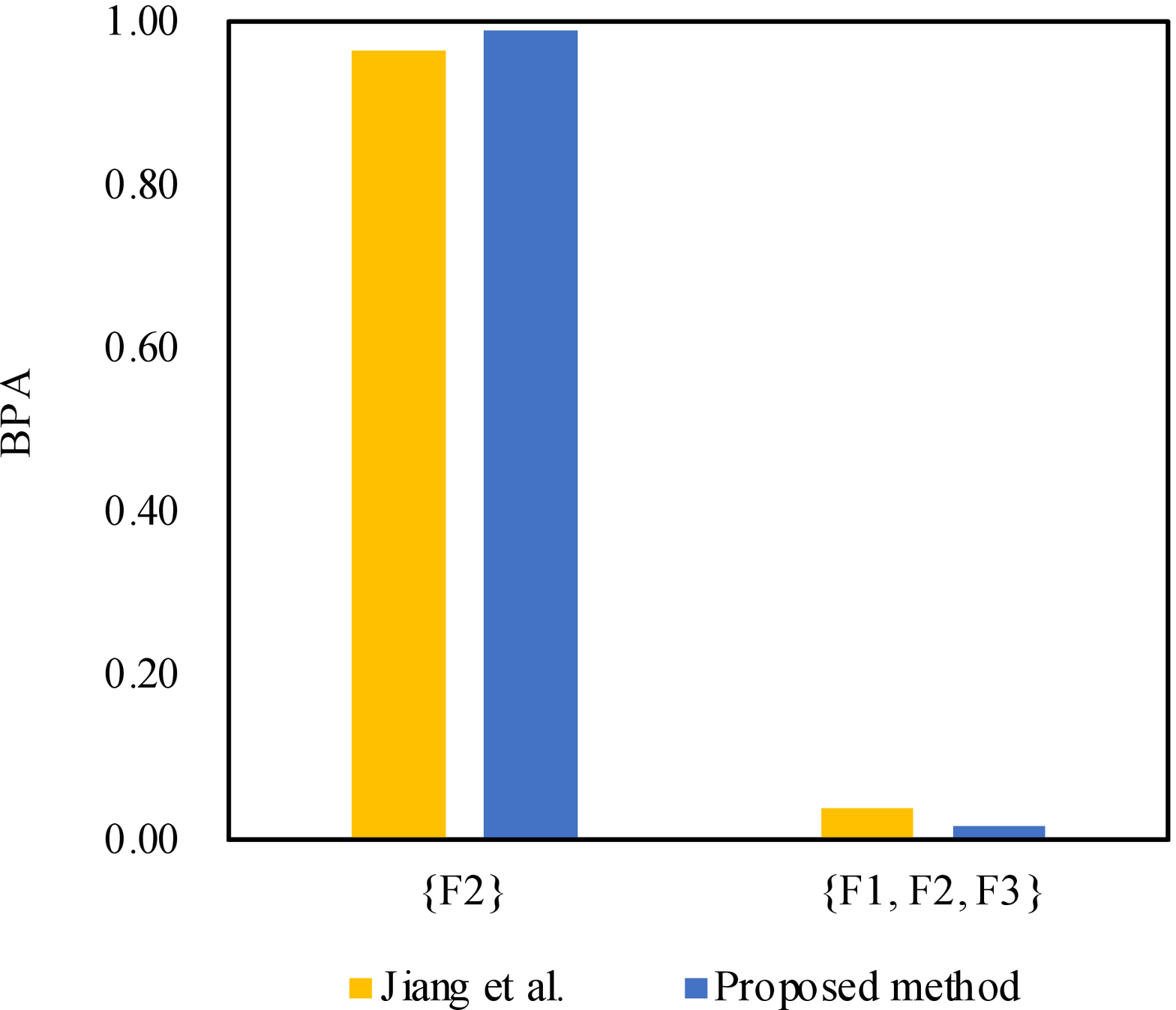}
}
\subfigure[]{
\label{exp4:b} %% label for second subfigure
\centering
\includegraphics[width=.475\linewidth,clip]{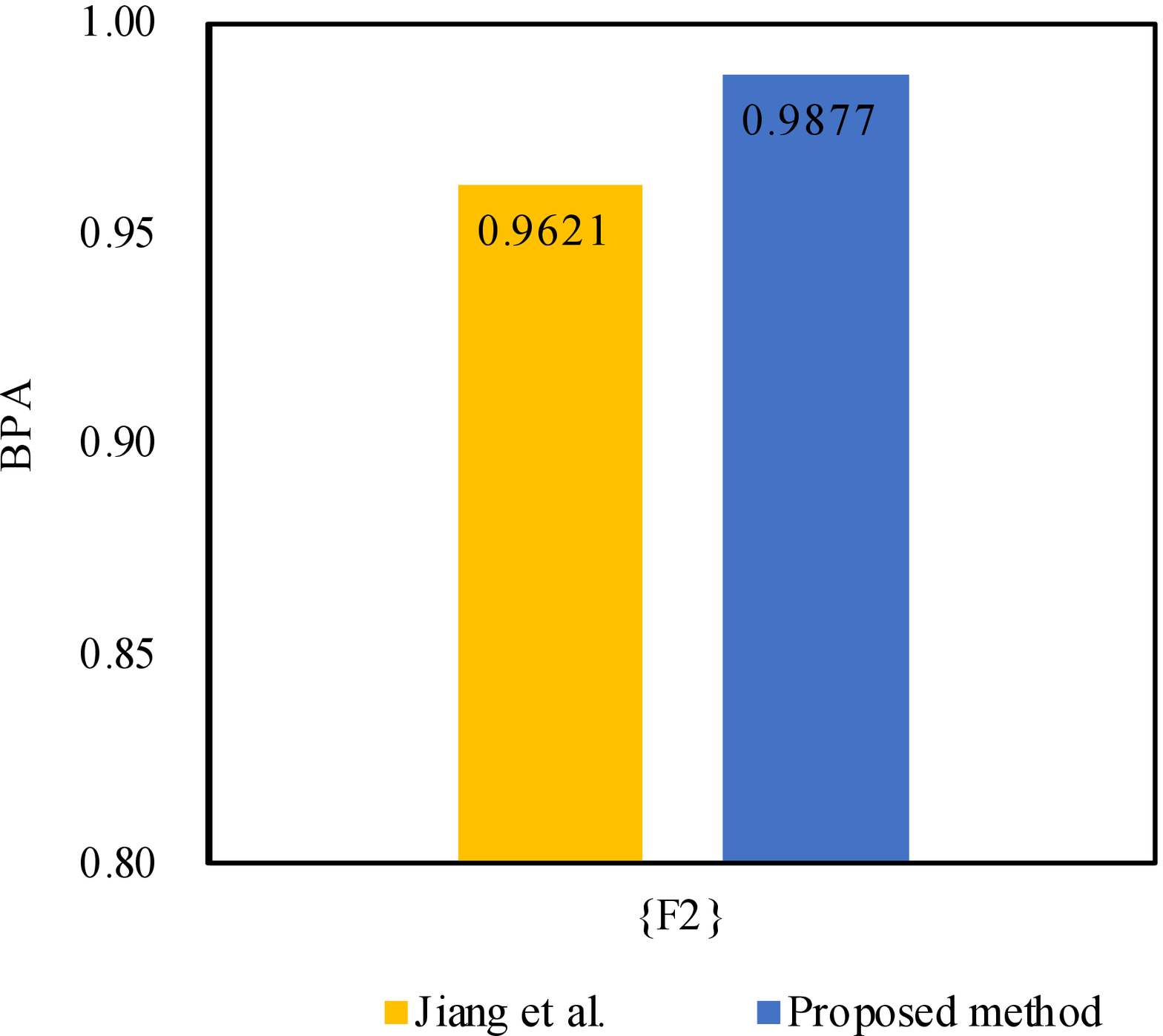}
}
\caption{The comparison of different methods for motor rotor fault diagnosis at $2X$ frequency.}
\label{2X} %% label for entire figure
\end{figure}

\begin{figure}
\centering
\subfigure[]{
\label{exp4:a} %% label for first subfigure
\includegraphics[width=.475\linewidth,clip]{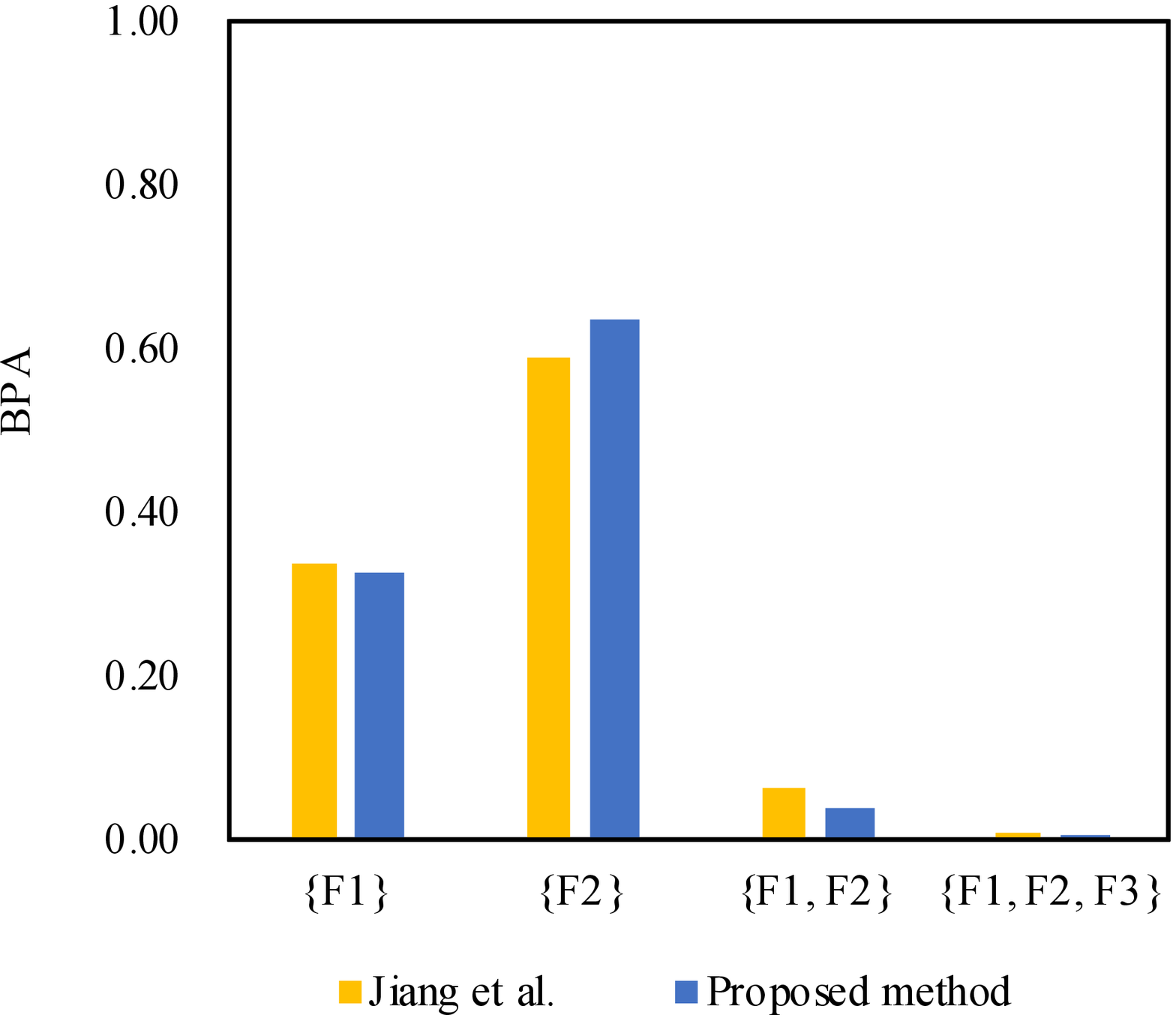}
}
\subfigure[]{
\label{exp4:b} %% label for second subfigure
\centering
\includegraphics[width=.475\linewidth,clip]{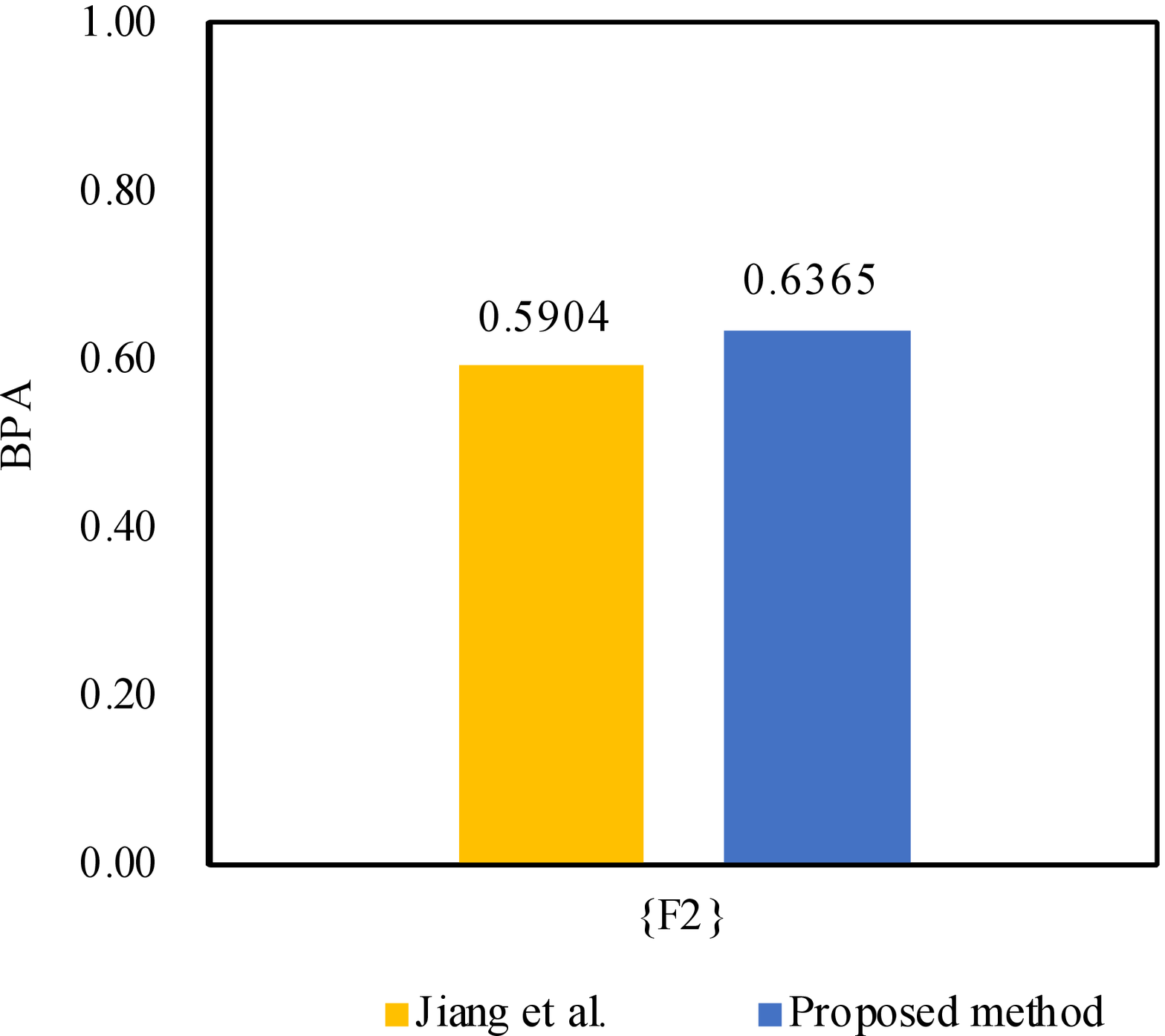}
}
\caption{The comparison of different methods for motor rotor fault diagnosis at $3X$ frequency.}
\label{3X} %% label for entire figure
\end{figure}

\subsection{Discussion}
From the results as shown in Tables~\ref{Fusionresults}, it can be noticed that the proposed method diagnoses the $F_2$ fault type, which is in accordance with Jiang et al.'s method~\cite{jiang2016sensordiagnosis}.
%Even facing the conflicting sensor reports where the normalized support degrees of the sensor reports are different at $1X$ frequency, $2X$ frequency and $3X$ frequency, both of the methods can well manage the conflicting pieces of evidence and diagnose the fault type $F_2$.

Additionally, as shown in Figures~\ref{1X}--\ref{3X}, the proposed method outperforms Jiang et al.'s method~\cite{jiang2016sensordiagnosis},
because by using Jiang et al.'s method, the belief degrees allocated to the target $F_2$ fault type at $1X$, $2X$ and $3X$ frequencies are 88.61\%, 96.21\% and 59.04\%, respectively, while the belief degrees allocated to the target $F_2$ fault type at $1X$, $2X$ and $3X$ frequencies by using the proposed method increase to 89.82\%, 98.77\% and 63.65\%, respectively.

\section{Conclusion}\label{Conclusion}
In this paper, by considering the discrepancy and conflict among evidences, a novel method for multi-sensor data fusion based on a new Generalised Belief Jensen-Shannon divergence measure method was proposed.
The proposed method consisted of three main steps.
Firstly, by taking into account the sufficiency and importance of the evidences, the reliability weight of each evidences was determined.
Next, based on the above reliability weights of evidences, the presented Generalised Belief Jensen-Shannon divergence was utilised to measure the support degrees of evidences, which was regarded as the final weight.
Afterwards, the final weights of the evidences were utilised to adjust the bodies of the evidences before using the Dempster's combination rule.
Finally, an application in fault diagnosis demonstrated that the proposed method could diagnose the faults more precisely.

%In the near future work, we intend to develop a generalised BJS divergence measure method to make it more applicable and efficient to fit the practical applications.
%Especially, considering those applications where different weights are assigned to decision makers, how can we develop an improved generalised BJS divergence measure method and apply it in reality will be investigated in the near future.

\section*{Conflict of Interest}
The author states that there are no conflicts of interest.

\section*{Acknowledgments}
%The author greatly appreciates the reviews' suggestions and the editor's encouragement.
This research is supported by the National Natural Science Foundation of China (Nos. 61672435, 61702427, 61702426) and the 1000-Plan of Chongqing by Southwest University (No. SWU116007).

\clearpage
\normalsize
\bibliographystyle{elsarticle-num}
%\section*{Reference}
%\bibliography{ref}

\end{document}